# Enhancing Convergence in Federated Learning: A Contribution-Aware Asynchronous Approach

Changxin Xu [1,*] , Yuxin Qiao [2] , Zhanxin Zhou [3] , Fanghao Ni [4] , Jize Xiong [5]

[1] Computer Information Technology, Northern Arizona University, Flagstaff, USA
[2] Big Data and Business Intelligence, Universidad Internacional Isabel I de Castilla, Burgos, Spain
[3] Computer Information Technology, Northern Arizona University, Flagstaff, USA
[4] Software Engineering, Indian Institute of Technology, Guwahati, Assam, India
[5] Computer Information Technology, Northern Arizona University, Flagstaff, USA

* **Corresponding author**: Farid Uddin Ahmed (Email: cx46@nau.edu)

**Abstract:** Federated Learning (FL) is a distributed machine learning paradigm that allows clients to train models on their data while preserving their privacy. FL algorithms, such as Federated Averaging (FedAvg) and its variants, have been shown to converge well in many scenarios. However, these methods require clients to upload their local updates to the server in a synchronous manner, which can be slow and unreliable in realistic FL settings. To address this issue, researchers have developed asynchronous FL methods that allow clients to continue training on their local data using a stale global model. However, most of these methods simply aggregate all of the received updates without considering their relative contributions, which can slow down convergence. In this paper, we propose a contribution-aware asynchronous FL method that takes into account the staleness and statistical heterogeneity of the received updates. Our method dynamically adjusts the contribution of each update based on these factors, which can speed up convergence compared to existing methods.

## 1. INTRODUCTION

Over the past few years, Machine Learning has become a hot research topic. Federated Learning (FL) is a privacy-preserving distributed machine learning paradigm where clients utilize their local private data to collaboratively train a global model under the coordination of a centralized server. Most existing FL methods are developed under a synchronous communication setting, in which clients can upload their local models to the server simultaneously. It implicitly requires the assumption that all selected clients can finish the local training steps at roughly the same time. This assumption, however, hardly holds in many real-world systems. The devices typically have different computation resources and communication bandwidths, making the local updates received by the server asynchronous. The devices that finish the local training and upload to the server lately are defined as stragglers. A naive method to compromise the scenario is to request the faster clients waiting for the stragglers to participate in the aggregation. Although this method can somehow balance the contributions of each device, it largely decays the convergence speed of the whole FL system and causes computation resource waste on the faster devices.

To overcome the straggler problem in synchronous FL, several asynchronous FL methods have been developed to improve the convergence performance. In the fully asynchronous FL methods, the server performs a global model update once it receives any one local information without any waiting. Although this kind of fully asynchronous method helps increase system efficiency, it could decay the global model performance since the training bias is introduced during each global update. In addition, too frequent global aggregations cause a waste of server resources. Therefore, has proposed another line of asynchronous FL, in which the server updates the global model once receiving the gradient from the first K clients' update while allowing the stragglers to continue their local training. Although this method achieves comparable performance with synchronous FL, it fails to measure the reasonable contributions of each local update during the aggregation step, as the server equally averages the received M local updates. The nature of different computation and communication abilities and statistical data heterogeneity cause the contributions of each received update different. In this work, we propose a systematic method to measure the contribution of each local update, aiming to address the following problems:

Problem 1: Statistical Heterogeneity. Statistical Heterogeneity (Non-IID) is considered as a challenging problem in federated learning. This problem is even accumulated in asynchronous federated learning scenarios. As devices communicate with the server under different frequencies, the global model would easily be biased to the faster devices. For each global model update, the weights of devices that contain the classes of data seldom studied should be higher than the devices with well-studied data.

Problem 2: Staleness. The heterogeneous computation and communication resources of each device create the problem of staleness. Without loss of the generality, assume for each client $i$ which contributes to the $t$-th server update, the local update uploaded is trained with global model version $t'$ and the staleness of this client $i$ is defined as $\tau_i(t) = t - t'$. Intuitively, the gradients with low stale are relatively more reliable, while the gradients with high stale are more risky to borrow biased information. Thus, the existing asynchronous methods downweight the contribution of gradient based on the staleness(e.g. $s(\tau_i(t)) = \frac{1}{(1+\tau_i(t))^{0.5}}$).

Although this kind of staleness-based weighted aggregation somehow reduces the risk of biased information, it could over-deteriorate the importance of high-stale gradients from two aspects. On the one hand, measuring contributions which depend on the delayed time to upload, can neglect the global model differences. For a simple example of a two-class classification task, we consider that there are two clients, which client 1 has data with class A and client 2 has data with class B. If client 2 last communicates with the server when the server has already learned enough information from class A and the global model has been approaching stable on class A, the global model would change little, nevertheless the staleness of client 2's communication. In this case, downweighting the contribution of client 2's gradient is meaningless. On the other hand, reducing the weight of the high-stale gradient further increases the risk of statistical training bias, as the slow devices could contain data that has been hardly studied.

## 2. RELATED WORK

### 2.1. Federated Learning

Federated Learning (FL) is a distributed machine learning framework that allows clients to collaboratively train a global model using their private data, under the coordination of a centralized server. Early works on FL, such as Federated Averaging (FedAvg), focused on reducing the communication overhead between the server and clients by using local stochastic gradient descent. Subsequent works have improved upon FedAvg by addressing the negative effects of statistical heterogeneity among clients. However, most of these works assume a synchronous communication setting, ignoring the system heterogeneity between clients. To address the issue of system heterogeneity, two types of methods have been proposed. One approach focuses on reducing the local computation workload of slower devices through partial model training in FL. In this paper, we focus on the other type of method, which allows faster and slower devices to communicate with the server asynchronously. This approach allows slower devices to continue training on their local data using a stale global model, without delaying the convergence of the overall FL system.

### 2.2. Asynchronous Federated Learning

Most existing works on asynchronous distributed learning, such as assume IID data distributions among the clients, which does not hold in the case of FL due to its inherent statistical heterogeneity. Other works in asynchronous federated learning, such as make additional assumptions, such as broadcasting model updates to all clients or assuming all clients have the same speed, to facilitate asynchronous FL.

One recent line of research on asynchronous FL updates the global model as soon as it receives any local information from clients, known as fully asynchronous FL. Although this kind of fully asynchronous FL helps to increase the system efficiency, it could enhance the cost of the server, as too frequent server aggregations are required. In addition, as the server aggregates each client's update, privacy could be a concern.

To solve the above problem, the most recent work studies another line of asynchronous FL, where the server updates the global model once receiving the gradient from the first $K$ client's updates. However, these works equally aggregate the received gradient and ignore the different contributions among the gradients. In this work, we propose a novel FL method that takes into account the contributions of each update and adjusts them accordingly. Our method uses a combination of weighting based on staleness and statistical heterogeneity to improve convergence performance.

## 3. MODEL AND PRELIMINARIES

We consider a FL system with one server and $N$ clients. The clients work together to train a machine learning model by solving a distributed optimization problem as follows:

$$\min_x f(x) = \frac{1}{N}\sum_{i=1}^{N} f_i(x) = \frac{1}{N}\sum_{i=1}^{N}\mathbb{E}_{\zeta_i}[F_i(x,\zeta_i)], \quad (1)$$

where $f_i: R^d \to R$ is a non-convex loss function for client $i$ and $F_i$ is the estimated loss function based on a mini-batch data sample $\zeta_i$ drawn from client $i$'s own dataset. Denote the dataset size of client $i$ as $D_i$.

The classic fully asynchronous FL methods update the global model once the server receives any local updates from the client. Although this kind of method shows comparable convergence speed with synchronous settings, it raises privacy concerns. Recently, proposes a new line of asynchronous FL, in which the server has a buffer to store the received local updates. The server will only update the global model when the buffer contains $K$ clients' local updates. (Note that the buffer size $K$ is a hyper-parameter.) This kind of method is suitable to combine with the secured aggregation methods to prevent clients' privacy. Thus, in this paper, we choose to follow the main structure of .

Let $\Delta^i_t$ be the local update uploaded by client $i$ to participate in the $t$ round global aggregation. $\Delta^i_t$ is the cumulative gradients of $M$ local training steps with initial model of global version $t - \tau_i^t$. In FedBuff, once the buffer collect $K$ local updates, it will aggregate the new global model $x_{t+1}$ as following:

$$x_{t+1} = x_x - \eta_g \frac{1}{K}\sum_{i=1}^{K}\Delta^i_t, \quad (2)$$

where $\eta_g$ is the global learning rate at the server side. Although FedBuff performs well under moderate conditions, its performance could highly drop as it ignores the difference among the contributions of $K$ local updates.

# 4. METHODOLOGY

The difference in contributions among *K* local updates comes from two aspects. On the first aspect, the system heterogeneity makes the required times to finish the *M* local training different. Typically the fast devices finish the local training sooner and upload timely information to the server, while the slow devices need more time to finish the local training and thus upload stale cumulative gradients to the server. On the other aspect, the statistical heterogeneity makes the different importance of clients' local updates to the current round model aggregation. In the following, we propose a novel method to measure the contribution of each client's local updates.

## 4.1. Staleness Effect

We here denote $\tau_i^t$ as the staleness between the model version in which local client *i* uses to start local training, and the model version in which client *i* participates in the *t* round global aggregation. Previous works down-weight stale updates using the function which is inversely propitiatory to $\tau_i^t$. (e.g., $s(\tau_i(t)) = \frac{1}{(1+\tau_i(t))^{0.5}}$). We measure the client *i*'s staleness degree $S_i^t$ in *t*-th global aggregation as following:

$$S_i^t = \frac{\min_{j \in \mathcal{K}} \|x^t - x^{t-\tau_j^t}\|^2}{\|x^t - x^{t-\tau_i^t}\|^2}, \quad (3)$$

where K is the set of local clients stored in the buffer to participate in the current round aggregation. Note that $S_i^t \in$. The client *i* that starts local training with the model version $x^{t-\tau_i^t}$ most similar to the current model version $x^t$ has the highest $S_i^t = 1$.

## 4.2. Statistical Effect

Although the staleness effect is an important component in asynchronous FL scenarios, only applying staleness when considering the local clients' contributions could induce more problems. For example, consider a slow device that contains the classes of data which has been rarely learned by the server, simply applying staleness degree $S_i^t$ could exacerbate the statistical heterogeneity. Thus, when we measure the contribution of clients' local updates, we should consider the clients' local statistical distributions. Unfortunately, privacy is a serious problem in the FL setting, which makes it impossible to directly report the local data distribution to the server. Moreover, as the statistical benefit of each client's local data to the server, keeps changing during each global model update, we require the 'fresh' information which is calculated based on the current global model version to determine the statistical effect. Contributes that in client selection, selecting the client with higher local loss could enhance the convergence speed. We propose to utilize the batch loss to estimate the local training loss. Typically, the higher training loss represents a worse learning of the current global model on local data. Thus, we define the statistical effect of client *i* in *t*-th global aggregation as:

$$P_i^t = \mathcal{N}_i \frac{1}{|\zeta_i|} F_i(x^t, \zeta_i) \quad (4)$$

where $\zeta_i$ is the mini-batch data sample randomly drawn from client *i*'s local dataset.

Then we can update the global model as follows:

$$x_{t+1} = x_x - \eta_g \frac{1}{K} \sum_{i=1}^{K} P_i^t \Delta_t^i / S_i^t, \quad (5)$$

where $\eta_g$ is the global learning rate at the server side.

# 5. EXPERIMENT

We conduct experiments for the image classification task on Fashion-MNIST [34] for non-IID data distributions. There are 30 clients and each client has 1500 instances for training. In each FL round, all clients participate. We use LeNet as the backbone model.

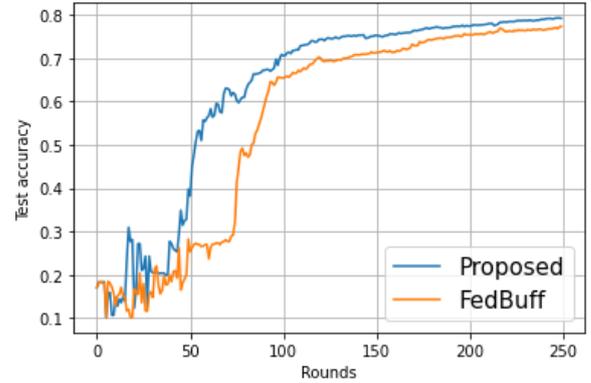

Fig. 1. Performance Comparison

The result shown in Fig. 1 validates that our proposed method outperforms the baseline method by a large margin.

# 6. CONCLUSION

In conclusion, this paper introduces a novel contributionaware asynchronous Federated Learning (FL) method that represents a significant advancement in distributed machine learning. By addressing the limitations of traditional synchronous FL methods, which often suffer from slow and unreliable communication in realistic settings, our approach offers a more efficient and practical solution. Unlike existing asynchronous FL methods that aggregate updates indiscriminately, our method considers the staleness and statistical heterogeneity of each update, dynamically adjusting their contributions to the global model. This nuanced approach ensures faster convergence and enhances the overall performance of FL systems. Our contribution-aware method not only mitigates the challenges posed by asynchronous updates but also paves the way for more robust and scalable FL implementations. Future research could further refine this approach, exploring additional factors that influence the effectiveness of updates and extending our method to a wider range of applications. Through continued innovation in FL, we can better harness the power of distributed machine learning while upholding the essential principles of privacy and efficiency.